\documentclass{article}

\usepackage[accepted]{icml2014}

\usepackage{times}

\usepackage{graphicx}
\usepackage{subfigure}

\usepackage{natbib} 
\usepackage{hyperref}
\usepackage{verbatim}
\usepackage{xfrac}

\icmltitlerunning{Transformations between Bayes Nets and Neural Nets}

\usepackage{amsthm}
\usepackage{amsmath}
\usepackage{amssymb}

\usepackage{tikz}
\usetikzlibrary{bayesnet}

\newcommand{\bb}[1]{\mathbf{#1}}
\newcommand{\bx}{\bb{x}}
\newcommand{\bxi}{\bx^{(i)}}
\newcommand{\bt}{\bb{t}}
\newcommand{\bT}{\boldsymbol{\theta}}
\newcommand{\by}{\bb{y}}
\newcommand{\bv}{\bb{v}}
\newcommand{\bw}{\bb{w}}

\newcommand{\beps}{\boldsymbol{\epsilon}}
\newcommand{\bz}{\bb{z}}
\newcommand{\bpa}{\bb{pa}}

\newcommand{\pT}{p_{\bT}}

\newcommand{\ep}{\epsilon}

\newcommand{\Exp}[2]{\mathbb{E}_{#1}\left[#2\right]}

\newcommand{\eqnr}{\addtocounter{equation}{1}\tag{\theequation}}

\newcommand{\Lz}{L^{(z)}}
\newcommand{\Lnz}{L^{(\setminus z)}}
\newcommand{\Lcz}{L^{(z\rightarrow)}}

\DeclareMathOperator*{\argmax}{argmax}

\newcommand{\fpd}[2]{\frac{\partial #1}{\partial #2}}
\newcommand{\spd}[3]{\frac{\partial^2 #1}{\partial #2 \partial #3}}

\theoremstyle{definition}

\begin{document} 

\twocolumn[
\icmltitle{Efficient Gradient-Based Inference through \texorpdfstring{\\}{} Transformations between Bayes Nets and Neural Nets}

\icmlauthor{Diederik P. Kingma}{d.p.kingma@uva.nl}
\icmlauthor{Max Welling}{m.welling@uva.nl}
\icmladdress{Machine Learning Group, University of Amsterdam}

\icmlkeywords{machine learning}

\vskip 0.3in
]

\begin{abstract}
Hierarchical Bayesian networks and neural networks with stochastic hidden units are commonly perceived as two separate types of models. We show that either of these types of models can often be transformed into an instance of the other, by switching between centered and differentiable non-centered parameterizations of the latent variables. The choice of parameterization greatly influences the efficiency of gradient-based posterior inference; we show that they are often complementary to eachother, we clarify when each parameterization is preferred and show how inference can be made robust. In the non-centered form, a simple Monte Carlo estimator of the marginal likelihood can be used for learning the parameters. Theoretical results are supported by experiments.

%

\end{abstract} 

\section{Introduction}

\emph{Bayesian networks} (also called \emph{belief networks}) are probabilistic graphical models where the conditional dependencies within a set of random variables are described by a directed acyclic graph (DAG). Many supervised and unsupervised models can be considered as special cases of Bayesian networks.

In this paper we focus on the problem of efficient inference in Bayesian networks with multiple layers of continuous latent variables, where exact posterior inference is intractable (e.g. the conditional dependencies between variables are nonlinear) but the joint distribution is differentiable. 
Algorithms for approximate inference in Bayesian networks can be roughly divided into two categories: sampling approaches and parametric approaches. Parametric approaches include Belief Propagation~\cite{pearl1982reverend} or the more recent Expectation Propagation (EP)~\cite{minka2001expectation}. When it is not reasonable or possible to make assumptions about the posterior (which is often the case), one needs to resort to sampling approaches such as Markov Chain Monte Carlo (MCMC)~\cite{neal1993probabilistic}.  In high-dimensional spaces, gradient-based samplers such as Hybrid Monte Carlo~\cite{duane1987hmc} and the recently proposed no-U-turn sampler~\cite{hoffman2011no} are known for their relatively fast mixing properties.
When just interested in finding a mode of the posterior, vanilla gradient-based optimization methods can be used. 
The alternative parameterizations suggested in this paper can dramatically improve the efficiency of any of these algorithms.

\subsection{Outline of the paper}

After reviewing background material in~\ref{sec:background}, we introduce a generally applicable differentiable reparameterization of continuous latent variables into a differentiable non-centered form in section~\ref{sec:auxform}. In section~\ref{sec:curvatures} we analyze the posterior dependencies in this reparameterized form.
Experimental results are shown in section~\ref{sec:experiments}.




\section{Background}
\label{sec:background}

\paragraph{Notation.} 
We use bold lower case (e.g. $\bx$ or $\by$) notation for random variables and instantiations (values) of random variables. We write $\pT(\bx|\by)$ and $\pT(\bx)$ to denote (conditional) probability density (PDF) or mass (PMF) functions of variables. With $\bT$ we denote the vector containing all parameters; each distribution in the network uses a subset of $\bT$'s elements. Sets of variables are capitalized and bold, matrices are capitalized and bold, and vectors are written in bold and lower case.

\subsection{Bayesian networks}\label{sec:bayesnets}

\begin{figure}[t]
\begin{center}
\begin{tabular}{cc}
\multicolumn{2}{c}{
\resizebox {\columnwidth} {!} {
\begin{tikzpicture}
\node[const] (z1) {$\cdots\,\,$};
\node[latent, right=of z1] (z2) {$\bz_j$};
\node[const, right=of z2] (x) {$\,\,\cdots$};
\edge {z1} {z2};
\edge {z2} {x};

\node[const, right=2.0 of x] (_z1) {$\cdots\,\,$};
\node[det, right=of _z1] (_z2) {$\bz_j$};
\node[latent, above=of _z2] (_e2) {$\beps_j$};
\node[const, right=of _z2] (_x) {$\,\,\cdots$};
\edge {_z1} {_z2};
\edge {_z2} {_x};
\edge {_e2} {_z2};
\end{tikzpicture}
}
} \\
\hspace{12.5 mm} (a) & \hspace{23 mm} (b) \\
\end{tabular}
\end{center}
\caption{
{\bf(a)} The centered parameterization (CP) of a latent variable $\bz_j$. {\bf(b)} 
The differentiable non-centered parameterization (DNCP) where we have introduced an auxiliary 'noise' variable $\beps_j \sim \pT(\beps_j)$ such that $\bz_j$ becomes deterministic: $\bz_j = g_j(\bpa_j, \beps_j, \bT)$. This deterministic variable has an interpretation of a hidden layer in a neural network, which can be differentiated efficiently using the backpropagation algorithm.
}\label{auxformfig}
\end{figure}
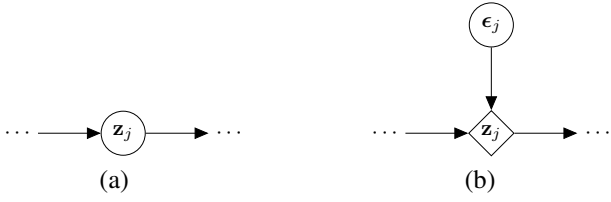

A Bayesian network models a set of random variables $\bb{V}$ and their conditional dependencies as a directed acyclic graph, where each variable corresponds to a vertex and each edge to a conditional dependency. Let the distribution of each variable $\bv_j$ be $\pT(\bv_j|\bpa_j)$, where we condition on $\bv_j$'s (possibly empty) set of parents $\bpa_j$. Given the factorization property of Bayesian networks, the joint distribution over all variables is simply:
\begin{align}\label{eq:bnjoint}
\pT(\bv_1, \dots, \bv_N) = \prod_{j=1}^N \pT(\bv_j|\bpa_j)
\end{align}

Let the graph consist of one or more (discrete or continuous) observed variables $\bx_j$ and continuous latent variables $\bz_j$, with corresponding conditional distributions $\pT(\bx_j|\bpa_j)$ and $\pT(\bz_j|\bpa_j)$. We focus on the case where both the marginal likelihood $\pT(\bx) = \int_\bz \pT(\bx, \bz) \,d\bz$ and the posterior $\pT(\bz|\bx)$ are intractable to compute or differentiate directly w.r.t. $\bT$ (which is true in general), and where the joint distribution $\pT(\bx, \bz)$ is at least once differentiable, so it is still possible to efficiently compute first-order partial derivatives $\nabla_{\bT} \log \pT(\bx, \bz)$ and $\nabla_{\bz} \log \pT(\bx, \bz)$. 

\subsection{Conditionally deterministic variables}\label{sec:detvars}
A \emph{conditionally deterministic variable} $\bv_j$ with parents $\bpa_j$ is a variable whose value is a (possibly nonlinear) deterministic function $g_j(.)$ of the parents and the parameters:
$\bv_j = g_j(\bpa_j, \bT)$.
The PDF of a conditionally deterministic variable is a Dirac delta function, which we define as a Gaussian PDF $\mathcal{N}(.;\mu,\sigma)$ with infinitesimal $\sigma$:
\begin{align}
\pT(\bv_j|\bb{pa}_j) = \lim_{\sigma \to 0} \mathcal{N}(\bv_j; g_j(\bpa_j, \bT), \sigma)
\label{diracPDF}
\end{align}
which equals $+\infty$ when $\bv_j = g_j(\bpa_j, \bT)$ and equals 0 everywhere else such that $\int_{\bv_j} \pT(\bv_j|\bb{pa}_j) \,d\bv_j = 1$. 

\subsection{Inference problem under consideration}\label{problem}
We are often interested in performing posterior inference, which most frequently consists of either optimization (finding a mode $\argmax_\bz \pT(\bz|\bx)$) or sampling from the posterior $\pT(\bz|\bx)$. Gradients of the log-posterior w.r.t. the latent variables can be easily acquired using the equality:
\begin{align*}
\nabla_{\bz} \log \pT(\bz|\bx) 
&= \nabla_{\bz} \log \pT(\bx,\bz) \\
&= \sum_{j=1}^N \nabla_{\bz} \log \pT(\bv_j | \bpa_j)\eqnr
\label{gradPosterior}
\end{align*}
In words, the gradient of the log-posterior w.r.t. the latent variables is simply the sum of gradients of individual factors w.r.t. the latent variables. These gradients can then be followed to a mode if one is interested in finding a MAP solution. If one is interested in sampling from the posterior then the gradients can be plugged into a gradient-based sampler such as Hybrid Monte Carlo~\cite{duane1987hmc}; if also interested in learning parameters, the resulting samples can be used for the \emph{E}-step in Monte Carlo EM~\cite{wei1990mcem} (MCEM).

Problems arise when strong posterior dependencies exist between latent variables. From eq.~\eqref{gradPosterior} we can see that the Hessian $\bb{H}$ of the posterior is:
\newcommand{\Hij}{\spd{\log \pT(\bz|\bx)}{z_i}{z_j}}
\begin{align*}
\bb{H}
&= \nabla_\bz \nabla_\bz^T \log \pT(\bz|\bx) = \sum_{j=1}^N \nabla_\bz \nabla_\bz^T \log \pT(\bv_j | \bpa_j)
\label{hessianPosterior}
\eqnr
\end{align*}
Suppose a factor $\log \pT(z_i | z_j)$ connecting two scalar latent variables $\bz_i$ and $\bz_j$ exists, and $z_i$ is strongly dependent on $z_j$, then the Hessian's corresponding element $\Hij$ will have a large (positive or negative) value. This is bad for gradient-based inference since it means that changes in $z_j$ have a large effect on the gradient $\fpd{\log \pT(z_i | z_j)}{z_i}$ and changes in $z_i$ have a large effect on the gradient $\fpd{\log \pT(z_i | z_j)}{z_j}$.
In general, strong conditional dependencies lead to ill-conditioning of the posterior, resulting in smaller optimal stepsizes for first-order gradient-based optimization or sampling methods, making inference less efficient.

\section{The differentiable non-centered parameterization (DNCP)}\label{sec:auxform}

In this section we introduce a generally applicable transformation between continuous latent random variables and deterministic units with auxiliary parent variables. In rest of the paper we analyze its ramifications for gradient-based inference.



\subsection{Parameterizations of latent variables}\label{sec:cp_and_dncp}

Let $\bz_j$ be some continuous latent variable with parents $\bpa_j$, and corresponding conditional PDF:
\begin{align*}
\bz_j | \bpa_j \sim \pT(\bz_j|\bpa_j)
\eqnr\label{eq:cp}\end{align*}
This is also known in the statistics literature as the \emph{centered parameterization} (CP) of the latent variable $\bz_j$. 
Let the \emph{differentiable non-centered parameterization} (DNCP)~ of the latent variable $\bz_j$ be:
\begin{align*}
\bz_j = g_j(\bpa_j, \beps_j, \bT) \text{\quad where\quad} \beps_j \sim p(\beps_j)
\eqnr\label{eq:dncp}
\end{align*}
where $g_j(.)$ is some differentiable function. 
Note that in the DNCP, the value of $\bz_j$ is \emph{deterministic} given both $\bpa_j$ and the newly introduced auxiliary variable $\beps_j$ which is distributed as $p(\beps_j)$. See figure~\ref{auxformfig} for an illustration of the two parameterizations.

By the change of variables, the relationship between the original PDF $\pT(\bz_j|\bpa_j)$, the function $g_j(\bpa_j, \beps_j)$ and the PDF $p(\beps_j)$ is:
\begin{align*}
p(\beps_j) = \pT(\bz_j = g_j(\bpa_j, \beps_j, \bT)|\bpa_j) \; |det(\bb{J})|
\eqnr\label{eq:changeofvariables}\end{align*} 
where $det(\bb{J})$ is the determinant of Jacobian of $g_j(.)$ w.r.t. $\beps_j$. If $z_j$ is a scalar variable, then $\epsilon_j$ is also scalar and $|det(\bb{J})| = |\fpd{z_j}{\epsilon_j}|$.

In the DNCP, the original latent variable $\bz_j$ has become deterministic, and its PDF $\pT(\bz_j | \bpa_j, \beps_j)$ can be described as a Dirac delta function (see section~\ref{sec:detvars}).


The joint PDF over the random and deterministic variables can be integrated w.r.t. the determinstic variables. If for simplicity we assume that observed variables are always leaf nodes of the network, and that all latent variables are reparameterized such that the only \emph{random} variables left are the observed and auxiliary variables $\bx$ and $\beps$, then the marginal joint $\pT(\bx, \beps)$ is obtained as follows:
\begin{align*}
&\pT(\bx, \beps) = \int_{\bz} \pT(\bx, \bz, \beps) \,d\bz\\
&= \int_{\bz}
\prod_j \pT(\bx_j|\bpa_j)
\prod_j \pT(\bz_j|\bpa_j, \beps_j)
\prod_j p(\beps_j)  \,d\bz
\\
&= \prod_j\pT(\bx_j|\bpa_j)
\prod_j p(\beps_j)
\int_{\bz} \prod_j \pT(\bz_j|\bpa_j, \beps_j) \,d\bz
\\
&= \prod_j \pT(\bx_j|\bpa_j)
\prod_j p(\beps_j)
\\
&\text{\hspace{5mm}where\hspace{5mm}$\bz_k = g_k(\bpa_k, \beps_k, \bT)$ } \\
\eqnr
\label{eq:integrating_out_z}
\end{align*}
In the last step of eq.~\eqref{eq:integrating_out_z}, the inputs $\bpa_j$ to the factors of observed variables $\pT(\bx_j|\bpa_j)$ are defined in terms of functions $\bz_k = g_k(.)$, whose values are all recursively computed from auxiliary variables $\beps$. 

\subsection{Approaches to DNCPs}
\label{sec:applicability}

There are a few basic approaches to transforming CP of a latent variable $\bz_j$ to a DNCP:
\begin{enumerate}
\item Tractable and differentiable inverse CDF. In this case, let $\epsilon_j \sim \mathcal{U}(0,1)$, and let $g_j(\bz_j,\bpa_j,\bT) = F^{-1}(\bz_j|\bpa_j;\bT)$ be the inverse CDF of the conditional distribution. Examples: Exponential, Cauchy, Logistic, Rayleigh, Pareto, Weibull, Reciprocal, Gompertz, Gumbel and Erlang distributions.

\item For any "location-scale" family of distributions (with differentiable log-PDF) we can choose the standard distribution (with $\text{location} =0$, $\text{scale} =1$) as the auxiliary variable $\beps_j$, and let $g_j(.)=\text{location}+\text{scale} \cdot \beps_j$. Examples: Gaussian, Uniform, Laplace, Elliptical, Student's t, Logistic and Triangular distributions.

\item Composition: It is often possible to express variables as functions of component variables with different distributions. Examples: Log-Normal (exponentiation of normally distributed variable), Gamma (a sum over exponentially distributed variables), Beta distribution, Chi-Squared, F distribution and Dirichlet distributions.
\end{enumerate}

When the distribution is not in the families above, accurate differentiable approximations to the inverse CDF can be constructed, e.g. based on polynomials, with time complexity comparable to the CP (see e.g. ~\cite{devroye1986sample} for some methods).

For the exact approaches above, the CP and DNCP forms have equal time complexities. In practice, the difference in CPU time depends on the relative complexity of computing derivatives of $\log \pT(\bz_j|\bpa_j)$ versus computing $g_j(.)$ and derivatives of $\log p(\epsilon_j)$, which can be easily verified to be similar in most cases below. Iterations with the DNCP form were slightly faster in our experiments.


\subsection{DNCP and neural networks}\label{sec:neuralnetwork}
It is instructive to interpret the DNCP form of latent variables as "hidden units" of a neural network. The network of hidden units together form a neural network with inserted noise $\beps$, which we can differentiate efficiently using the backpropagation algorithm~\cite{rumelhart1986backprop}.

There has been recent increase in popularity of deep neural networks with stochastic hidden units (e.g.~\cite{krizhevsky2012imagenet, goodfellow2013maxout,bengio2013estimating}). Often, the parameters $\bT$ of such neural networks are optimized towards maximum-likelihood objectives. In that case, the neural network can be interpreted as a probabilistic model $\log \pT(\bt|\bx,\beps)$ computing a conditional distribution over some target variable $\bt$ (e.g. classes) given some input $\bx$. In~\cite{bengio2013deep}, stochastic hidden units are used for learning the parameters of a Markov chain transition operator that samples from the data distribution.

For example, in ~\cite{hinton2012improving} a 'dropout' regularization method is introduced where (in its basic version) the activation of hidden units $z_j$ is computed as $z_j = \epsilon_j \cdot f(\bpa_j)$ with $\epsilon_j \sim p(\epsilon_j) =\text{Bernoulli}(0.5)$, and where the parameters are learned by following the gradient
of the log-likelihood lower bound: $\nabla_{\bT} \Exp{\beps}{ \log \pT(\bt^{(i)}|\bxi,\beps)}$; this gradient can sometimes be computed exactly~\cite{maaten2013learning} and can otherwise be approximated with a Monte Carlo estimate~\cite{hinton2012improving}. The two parameterizations explained in section~\ref{sec:cp_and_dncp} offer us a useful new perspective on 'dropout'. A 'dropout' hidden unit (together with its injected noise $\beps$) can be seen as the DNCP of latent random variables, whose CP is $z_j | \bpa_j \sim \pT(\bz_j = \epsilon_j \cdot f(\bpa_j) | \bpa_j))$. A practical implication is that  'dropout'-type neural networks can therefore be interpreted and treated as hierarchical Bayes nets, which opens the door to alternative approaches to learning the parameters, such as Monte Carlo EM or variational methods. 

While 'dropout' is designed as a regularization method, other work on stochastic neural networks exploit the power of stochastic hidden units for generative modeling, e.g.~\cite{frey1999variational, rezende2014stochastic, tang2013learning} applying (partially) MCMC or (partically) factorized variational approaches to modelling the posterior. As we will see in sections~\ref{sec:curvatures} and~\ref{sec:experiments}, the choice of parameterization has a large impact on the posterior dependencies and the efficiency of posterior inference. However, current publications lack a good justification for their choice of parameterization. The analysis in section~\ref{sec:curvatures} offers some important insight in where the centered or non-centered parameterizations of such networks are more appropriate.

\subsection{A differentiable MC likelihood estimator}
\label{sec:mclikelihood}
We showed that many hierarchical continuous latent-variable models can be transformed into a DNCP $\pT(\bx, \beps)$, where all latent variables (the introduced auxiliary variables $\beps$) are root nodes (see eq.~\eqref{eq:integrating_out_z}). This has an important implication for learning since (contrary to a CP) the DNCP can be used to form a differentiable Monte Carlo estimator of the marginal likelihood:
\begin{align*}
\log \pT(\bx)
&\simeq \log \frac{1}{L} \sum_{l=1}^L \prod_j \pT(\bx_j|\bpa_j^{(l)})
\end{align*}
where the parents $\bpa_j^{(l)}$ of the observed variables are either root nodes or functions of root nodes whose values are sampled from their marginal: $\beps^{(l)} \sim p(\beps)$. This MC estimator can be differentiated w.r.t. $\bT$ to obtain an MC estimate of the log-likelihood gradient $\nabla_{\bT} \log \pT(\bx)$, which can be plugged into stochastic optimization methods such as Adagrad for approximate ML or MAP. When performed one datapoint at a time, we arrive at our on-line Maximum Monte Carlo Likelihood (MMCL) algorithm.


\begin{figure}[t]
\begin{center}
\begin{tabular}{cc}
\multicolumn{2}{c}{
\resizebox {0.95\columnwidth} {!} {
\begin{tikzpicture}
\node[obs] (x1) {$\bx_1$};
\node[obs, right=of x1] (x2) {$\bx_2$};
\node[obs, right=of x2] (x3) {$\bx_3$};
\node[latent, above=of x1] (z1) {$\bz_1$};
\node[latent, above=of x2] (z2) {$\bz_2$};
\node[latent, above=of x3] (z3) {$\bz_3$};
\edge {z1} {z2};
\edge {z1} {x1};
\edge {z2} {x2};
\edge {z2} {z3};
\edge {z3} {x3};

\node[obs, right=2.0 of x3] (_x1) {$\bx_1$};
\node[obs, right=of _x1] (_x2) {$\bx_2$};
\node[obs, right=of _x2] (_x3) {$\bx_3$};
\node[det, above=of _x1] (_z1) {$\bz_1$};
\node[det, above=of _x2] (_z2) {$\bz_2$};
\node[det, above=of _x3] (_z3) {$\bz_3$};
\node[latent, above=of _z1] (_e1) {$\beps_1$};
\node[latent, above=of _z2] (_e2) {$\beps_2$};
\node[latent, above=of _z3] (_e3) {$\beps_3$};

\edge {_e1} {_z1};
\edge {_e2} {_z2};
\edge {_e3} {_z3};
\edge {_z1} {_z2};
\edge {_z2} {_z3};
\edge {_z1} {_x1};
\edge {_z2} {_x2};
\edge {_z3} {_x3};

\end{tikzpicture}
}
} \\
\hspace{13.5 mm} (a) & \hspace{23 mm} (b) \\
\end{tabular}
\end{center}
\caption{
{\bf(a)} An illustrative hierarchical model in its centered parameterization (CP). {\bf(b)} The differentiable non-centered parameterization (DNCP), where $\bz_1 = g_1(\beps_1, \bT)$, $\bz_2 = g_2(\bz_1, \beps_2, \bT)$ and $\bz_3 = g_3(\bz_2, \beps_3, \bT)$, with auxiliary latent variables $\beps_k \sim \pT(\beps_k)$. The DNCP exposes a neural network within the hierarchical model, which we can differentiate efficiently using backpropagation.
}\label{fig:auxformexample}
\end{figure}
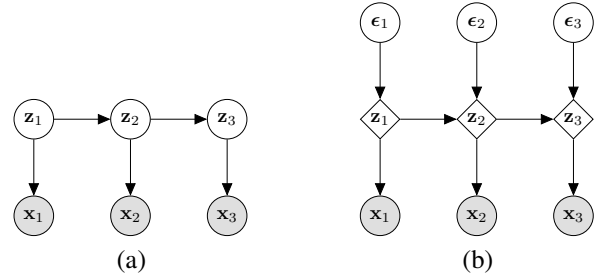

\section{Effects of parameterizations on posterior dependencies}\label{sec:curvatures}

\begin{figure}[t]
\centering
\includegraphics[width=1\columnwidth]{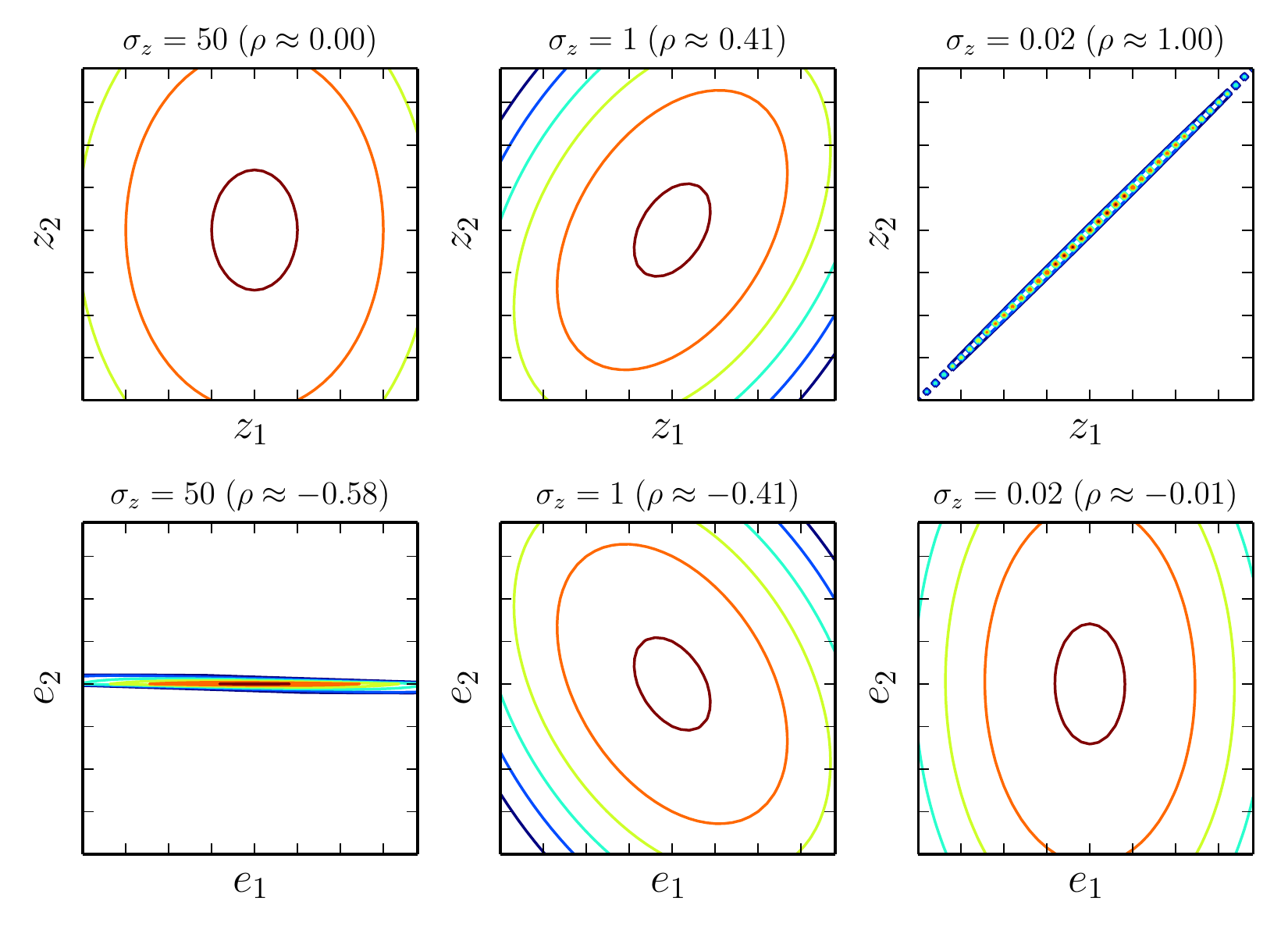}
\caption{Plots of the log-posteriors of the illustrative linear-Gaussian model discussed in sec.~\ref{sec:example}. Columns: different choices of $\sigma_z$, ranging from a low prior dependency ($\sigma_z=50$) to a high prior dependency ($\sigma_z=0.02$). First row: CP form. Second row: DNCP form. The posterior correlation $\rho$ between the variables is also displayed. In the original form a larger prior dependency leads to a larger posterior dependency (see top row). The dependency in the \emph{DNCP} posterior is inversely related to the prior dependency between $z_1$ and $z_2$ (bottom row).}
\label{fig:example_joints}
\end{figure}

\begin{table}[t]
\caption{Limiting behaviour of squared correlations between $z$ and its parent $y_i$ when $z$ is in the centered (CP) and non-centered (DNCP) parameterizaton.}
\label{table:limiting_behaviour}
\vskip 0.15in
\begin{center}
\begin{small}
\begin{sc}
\begin{tabular}[c]{lcc}
\hline
\abovespace\belowspace
 & 
$\rho^2_{y_i,z}$ (CP) &
$\rho^2_{y_i,e}$ (DNCP)
\\
\hline
\abovespace
$\lim_{\sigma \to 0}$ &
1 &
0 \\
$\lim_{\sigma \to +\infty}$ &
0 &
$\frac{\beta w_i^2}{\beta w_i^2 + \alpha}$ \\
$\lim_{\beta \to 0}$ &
$\frac{w_i^2}{w_i^2 - \alpha \sigma^2}$ &
0 \\
$\lim_{\beta \to -\infty}$ &
0 &
1 \\
$\lim_{\alpha \to 0}$ &
$\frac{1}{1 - \beta \sigma^2}$ &
$\frac{\beta \sigma^2}{\beta \sigma^2 - 1}$ \\
\belowspace
$\lim_{\alpha \to -\infty}$ &
0 &
0 \\
\hline \\
\end{tabular}
\end{sc}
\end{small}
\end{center}
\vskip -0.1in
\end{table}

What is the effect of the proposed reparameterization on the efficiency of inference?  If the latent variables have linear-Gaussian conditional distributions, we can use the metric of squared correlation between the latent variable and any of its children in their posterior distribution. If after reparameterization the squared correlation is decreased, then in general this will also result in more efficient inference.

For non-linear Gaussian conditional distributions, the log-PDF can be locally approximated as a linear-Gaussian using a second-order Taylor expansion. Results derived for the linear case can therefore also be applied to the non-linear case; the correlation computed using this approximation is a local dependency between the two variables.

Denote by $z$ a scalar latent variable we are going to reparameterize, and by $\by$ its parents, where $y_i$ is one of the parents. The log-PDF of the corresponding conditional distribution is
\begin{align*}
\log \pT(z|\by) &= \log \mathcal{N}(z|\bw^T \by + b, \sigma^2)\\
 &= - (z - \bw^T \by - b)^2/(2\sigma^2) + \text{constant}
\end{align*}
A reparameterization of $z$ using an auxiliary variable $\ep$ is $z = g(.) = (\bw^T \by + b) + \sigma \ep$ where $\ep \sim \mathcal{N}(0,1)$.
With~\eqref{eq:changeofvariables} it can be confirmed that this change of variables is correct:
\begin{align*}
\pT(z | \by) \cdot \left| \fpd{z}{\ep} \right|
&= \pT(z = g(.) | \by) \cdot \left| \fpd{z}{\ep} \right| \\
&= \mathcal{N}(\bw^T\by + b + \sigma \ep | \bw^T \by + b, \sigma^2) \cdot \sigma_z \\
&= - \exp(\ep^2/2) / \sqrt{2 \pi} = \mathcal{N}(0, 1) \\
&= p(\ep) 
\eqnr\label{eq:gaussian_reparam_correct}\end{align*}

First we will derive expressions for the squared correlations between $z$ and its parents, for the CP and DNCP case, and subsequently show how they relate.

The covariance $C$ between two jointly Gaussian distributed variables $A$ and $B$ equals the negative inverse of the Hessian matrix of the log-joint PDF:
\begin{align*}
C 
= \left( \begin{matrix}\sigma_A^2&\sigma_{AB}^2\\ \sigma_{AB}^2 & \sigma_{B}^2 \end{matrix} \right)
= - \bb{H}^{-1} 
= \frac{1}{det(\bb{H})} \left( \begin{matrix}-H_{B}&H_{AB}\\H_{AB}&-H_{A}\end{matrix} \right)
\end{align*}
The correlation $\rho$ between two jointly Gaussian distributed variables $A$ and $B$ is given by: $\rho = \sigma_{AB}^2 / (\sigma_A \sigma_B)$. Using the equation above,  the squared correlation can be computed from the elements of the Hessian matrix:
\begin{align*}
\rho^2 
&= (\sigma_{AB}^2)^2 / ( \sigma_A^2 \sigma_B^2 ) \\
&= (H_{AB}/det(\bb{H}))^2 / ((-H_{A}/det(\bb{H}))(-H_{B}/det(\bb{H})) \\
&= H_{AB}^2 / (H_{A} H_{B})
\eqnr \label{eq:correlation}
\end{align*}

Important to note is that derivatives of the log-posterior w.r.t. the latent variables are equal to the derivatives of log-joint w.r.t. the latent variables, therefore, 
\begin{align*}
\bb{H} = \nabla_{\bz} \nabla_{\bz}^T \log \pT(\bz|\bx) = \nabla_{\bz} \nabla_{\bz}^T \log \pT(\bx,\bz)
\end{align*}

The following shorthand notation is used in this section:
\begin{align*}
L &= \log \pT(\bx,\bz) \quad\text{(sum of all factors)} \\
z &= \text{the variable to be reparameterized}\\
\by &= \text{$z$'s parents}\\
\Lz &= \log \pT(z|\by) \quad\text{($z$'s factor)}\\
\Lnz &= L - \Lz \quad\text{(all factors minus $z$'s factor)}\\
\Lcz &= \text{the factors of $z$'s children}\\
\alpha &= \spd{\Lnz}{y_i}{y_i}\\
\beta &= \spd{\Lcz}{z}{z}\\
\end{align*}

\subsection{Squared Correlations}
\subsubsection{Centered case}
In the CP case, the relevant Hessian elements are as follows:
\begin{align*}
H_{y_i y_i} &= \spd{L}{y_i}{y_i} = \alpha + \spd{\Lz}{y_i}{y_i} = \alpha - w_i^2/ \sigma^{2} \\
H_{zz} &= \spd{L}{z}{z} = \beta + \spd{\Lz}{z}{z} = \beta - 1/\sigma^2 \\
H_{y_i z} &= \spd{L}{y_i}{z} = \spd{\Lz}{y_i}{z} = w_i/\sigma^2
\eqnr\end{align*}
Therefore, using eq.~\eqref{eq:correlation}, the squared correlation between $y_i$ and $z$ is:
\begin{align*}
\rho^2_{y_i,z} = \frac{(H_{y_i z})^2 }{H_{y_i y_i} H_{zz}}
= \frac{w_i^2/\sigma^4}{(\alpha - w_i^2/ \sigma^{2})(\beta - 1/\sigma^2)}
\eqnr\label{eq:corr_original}
\end{align*}

\subsubsection{Non-centered case}

In the DNCP case, the Hessian elements are:
\begin{align*}
H_{y_i y_i} &= \spd{L}{y_i}{y_i} = \alpha + \fpd{}{y_i} \fpd{\Lcz}{y_i} \\
&= \alpha + \fpd{}{y_i} \left( w_i \fpd{\Lcz}{z} \right)  =  \alpha  + w_i^2 \beta \\
H_{\ep \ep} &= \spd{L}{\ep}{\ep} = \spd{\Lcz}{\ep}{\ep} + \spd{\log p(\beps)}{\ep}{\ep} = \sigma^2 \beta - 1\\
H_{y_i \ep} &= \spd{L}{y_i}{\ep} = \sigma w_i \beta
\eqnr
\end{align*}
The squared correlation between $y_i$ and $\ep$ is therefore:
\begin{align*}
\rho^2_{y_i,\ep} 
= \frac{(H_{y_i \ep})^2 }{H_{y_i y_i} H_{\ep \ep}} 
&= \frac{\sigma^2 w_i^2 \beta^2}{(\alpha  + w_i^2 \beta)(\sigma^2 \beta - 1)}
\eqnr\label{eq:corr_reparameterized}
\end{align*}

\subsection{Correlation inequality}
We can now compare the squared correlation, between $z$ and some parent $y_i$, before and after the reparameterization. Assuming $\alpha < 0$ and $\beta < 0$ (i.e. $\Lnz$ and $\Lcz$ are concave, e.g. exponential families):
\begin{align*}
\rho^2_{y_i,z} &> \rho^2_{y_i,\ep} \\ 
\frac{w_i^2/\sigma^4}{(\alpha - w_i^2/ \sigma^{2})(\beta - 1/\sigma^2)}
&>
\frac{\sigma^2 w_i^2 \beta^2}{(\alpha  + w_i^2 \beta)(\sigma^2 \beta - 1)}
\\
\frac{w_i^2/\sigma^4}{(\alpha - w_i^2/ \sigma^{2})(\beta - 1/\sigma^2)}
&>
\frac{w_i^2 \beta^2}{(\alpha  + w_i^2 \beta)(\beta - 1/\sigma^2)}
\\
\frac{1/\sigma^4}{(\alpha - w_i^2/ \sigma^{2})}
&>
\frac{\beta^2}{(\alpha  + w_i^2 \beta)}
\\
\sigma^{-2} &> - \beta
\\
\eqnr\label{eq:corr_inequality}\end{align*}
Thus we have shown the surprising fact that the correlation inequality takes on an extremely simple form where the parent-dependent values $\alpha$ and $w_i$ play no role; the inequality only depends on two properties of $z$: the relative strenghts of $\sigma$ (its noisiness) and $\beta$ (its influence on children's factors).
Informally speaking, if the noisiness of $z$'s conditional distribution is large enough compared to other factors' dependencies on $z$, then the reparameterized form is beneficial for inference.

\subsection{A beauty-and-beast pair}
\label{sec:beauty_beast_pair}
Additional insight into the properties of the CP and DNCP can be gained by taking the limits of the squared correlations~\eqref{eq:corr_original} and~\eqref{eq:corr_reparameterized}. Limiting behaviour of these correlations is shown in table~\ref{table:limiting_behaviour}. As becomes clear in these limits, the CP and DNCP often form a beauty-and-beast pair: when posterior correlations are high in one parameterization, they are low in the other. This is especially true in the limits of $\sigma \to 0$ and $\beta \to -\infty$, where squared correlations converge to either $0$ or $1$, such that posterior inference will be extremely inefficient in either CP or DNCP, but efficient in the other. This difference in shapes of the log-posterior is illustrated in figure~\ref{fig:example_joints}.

\begin{table}[t]
\caption{Effective Sample Size (ESS) for different choices of latent-variable variance $\sigma_z$, and for different samplers, after taking 4000 samples. Shown are the results for HMC samplers using the \emph{CP} and \emph{DNCP} parameterizations, as well as a robust HMC sampler.}
\label{table:ess_logz}
\vskip 0.15in
\begin{center}
\begin{small}
\begin{sc}
\begin{tabular}[c]{lccc}
\hline
\abovespace\belowspace
$\log \sigma_z$ & 
CP &
DNCP &
Robust
\\
\hline
\abovespace
-5 &  2  &  305 & 640 \\
-4.5 & 26 &  348 & 498\\
-4  &10   &570  &686\\
-3.5    &225  &417  &624\\
-3  &386  &569  &596\\
-2.5    &542  &608  &900\\
-2  &406  &972  &935\\
-1.5    &672  &1078 &918\\
-1  &1460 &1600 &1082 \\
\hline \\
\end{tabular}
\end{sc}
\end{small}
\end{center}
\vskip -0.1in
\end{table}

\begin{figure}[t]
\begin{subfigure}[Centered Parameterization (CP)]{\includegraphics[width=7cm]{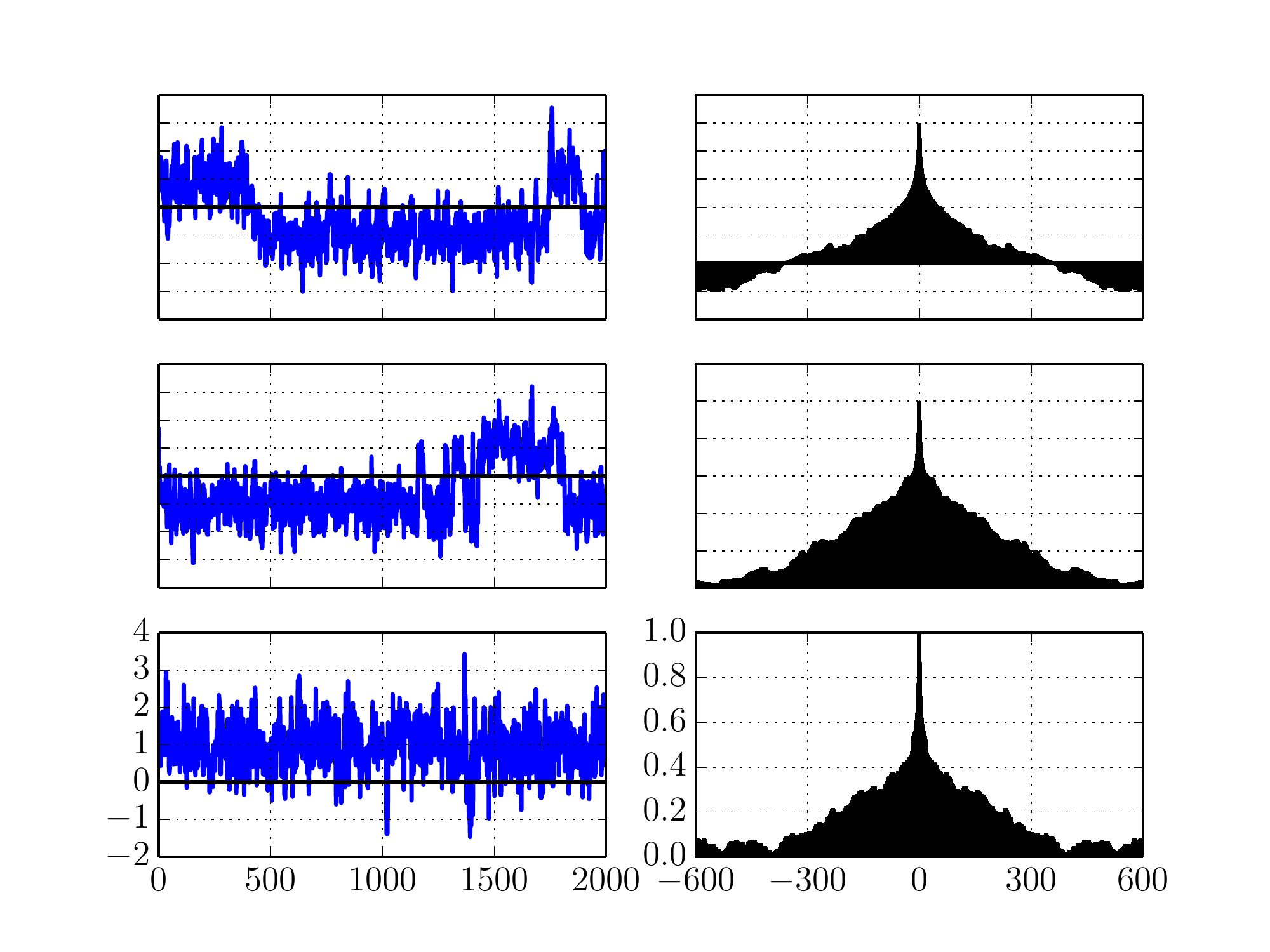}}
\end{subfigure}
\begin{subfigure}[Differentiable Non-Centered Parameterization (DNCP)]{\includegraphics[width=7cm]{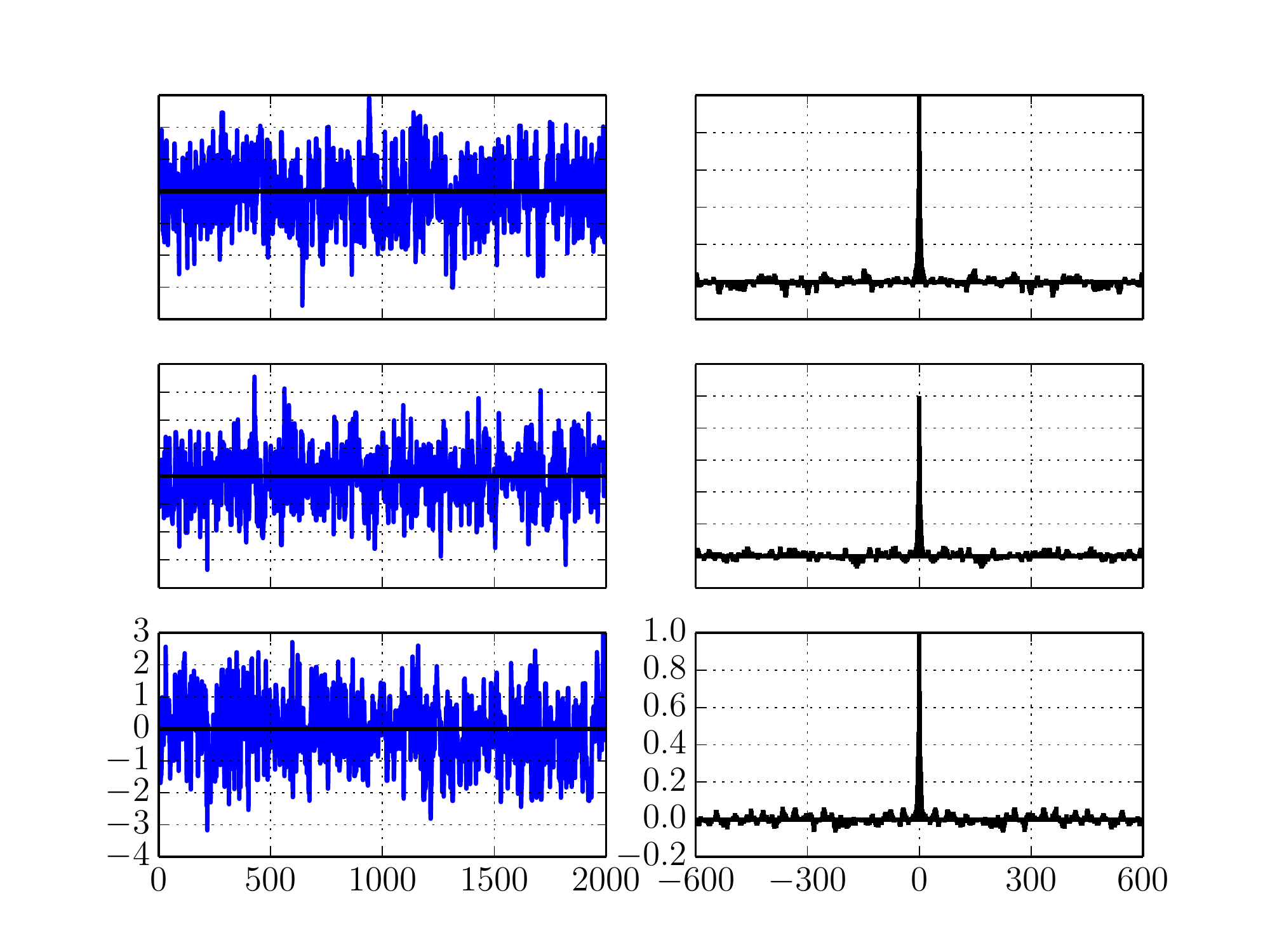}}
\end{subfigure}
\caption{Auto-correlation of HMC samples of the latent variables for a DBN in two different parameterizations. Left on each figure are shown 2000 subsequent HMC samples of three randomly chosen variables in the dynamic Bayesian network model. On the right are shown the corresponding HMC sample auto-correlation graphs. The DNCP resulted in much lower posterior dependencies and a dramatic drop in HMC sample auto-correlation.}
\label{fig:autocorr}
\end{figure}

\subsection{Example: Simple Linear Dynamical System}\label{sec:example}


Take a simple model with scalar latent variables $z_1$ and $z_2$, and scalar observed variables $x_1$ and $x_2$. The joint PDF is defined as $p(x_1, x_2, z_1, z_2) = p(z_1) p(x_1|z_1) p(z_2|z_1) p(x_2|z_2)$, where $p(z_1) = \mathcal{N}(0,1)$, $p(x_1|z_1)=\mathcal{N}(z_1,\sigma_x^2)$, $p(z_2|z_1)=\mathcal{N}(z_1,\sigma_z^2)$ and $p(x_2|z_2)=\mathcal{N}(z_2,\sigma_x^2)$. Note that the parameter $\sigma_z$ determines the dependency between the latent variables, and $\sigma_x$ determines the dependency between latent and observed variables.

We reparameterize $z_2$ such that it is conditionally deterministic given a new auxiliary variable $\ep_2$. Let $p(\ep_2) = \mathcal{N}(0,1)$. let $z_2 = g_2(z_1, \ep_2, \sigma_z) = z_1 + \sigma_z \cdot \ep_2$ and let $\ep_1 = z_1$. See figure~\ref{fig:example_joints} for plots of the original and auxiliary posterior log-PDFs, for different choices of $\sigma_z$, along with the resulting posterior correlation $\rho$.

For what choice of parameters does the reparameterization yield smaller posterior correlation? We use equation~\eqref{eq:corr_inequality} and plug in $\sigma \gets \sigma_z$ and $-\beta \gets \sigma^{-2}_x$, which results in:
\begin{align*}
\rho_{z_1, z_2}^2 > \rho_{\epsilon_1, \epsilon_2}^2 \quad\Rightarrow\quad
\sigma^2_z < \sigma^2_x
\end{align*}
i.e. the posterior correlation in DNCP form $\rho_{\epsilon_1, \epsilon_2}^2$ is smaller when the latent-variable noise parameter $\sigma^2_z$ is smaller than the oberved-variable noise parameter $\sigma^2_x$. Less formally, this means that the DNCP is preferred when the latent variable is more strongly coupled to the data (likelihood) then to its parents.


\section{Related work}
\label{sec:relatedwork}
This is, to the best of our knowledge, the first work to investigate the implications of the different differentiable non-centered parameterizations on the efficiency of gradient-based inference.
However, the topic of centered vs non-centered parameterizations has been investigated for efficient (non-gradient based) Gibbs Sampling in work by Papaspiliopoulos et~al.~\yrcite{papaspiliopoulos2003non, papaspiliopoulos2007general}, which also discusses some strategies for constructing parameterization for those cases. There have been some publications for parameterizations of specific models; ~\cite{gelfand1995efficient}, for example, discusses parameterizations of mixed models, and ~\cite{meng1998fast} investigate several rules for choosing an appropriate parameterization for mixed-effects models for faster EM.
In the special case where Gibbs sampling is tractable, efficient sampling is possible by interleaving between centered and non-centered parameterizations, as was shown in ~\cite{Yu11}.

Auxiliary variables are used for data augmentation (see \cite{van2001art} or slice sampling~\cite{neal2003slice}) where, in contrast with our method, sampling is performed in a higher-dimensional augmented space. Auxiliary variables are used in a similar form under the name exogenous variables in Structural Causal Models (SCMs)~\cite{pearl2000causality}. In SCMs the functional form of exogenous variables is more restricted than our auxiliary variables.
The concept of conditionally deterministic variables has been used earlier in e.g.~\cite{cobb2005nonlinear}, although not as a tool for efficient inference in general Bayesian networks with continuous latent variables.
%
%
Recently, \cite{raiko2012deep} analyzed the elements of the Hessian w.r.t. the parameters in neural network context.

The differentiable reparameterization of latent variables in this paper was introduced earlier in~\cite{kingma2013fast} and independently in \cite{bengio2013estimating}, but these publications lack a theoretic analysis of the impact on the efficiency of inference. In~\cite{kingma2013auto}, the reparameterization trick was used in an efficient algorithm for stochastic variational inference and learning.

\section{Experiments}
\label{sec:experiments}

\subsection{Nonlinear DBN}

From the derived posterior correlations in the previous sections we can conclude that depending on the parameters of the model, posterior sampling can be extremely inefficient in one parameterization while it is efficient in the other. When the parameters are known, one can choose the best parameterization (w.r.t. posterior correlations) based on the correlation inequality~\eqref{eq:corr_inequality}. 

In practice, model parameters are often subject to change, e.g. when optimizing the parameters with Monte Carlo EM; in these situations where there is uncertainty over the value of the model parameters, it is impossible to choose the best parameterization in advance. The "beauty-beast" duality from section~\ref{sec:beauty_beast_pair} suggests a solution in the form of a very simple sampling strategy: mix the two parameterizations. Let $Q_{CP}(\bz'|\bz)$ be the MCMC/HMC proposal distribution based on  $\pT(\bz|\bx)$ (the CP), and let $Q_{DNCP}(\bz'|\bz)$ be the proposal distribution based on $\pT(\beps|\bx)$ (the DNCP). Then the new MCMC proposal distribution based on the mixture is:
\begin{align*}
Q(\bz'|\bz) = \rho \cdot Q_{CP}(\bz'|\bz) + (1-\rho) \cdot Q_{DNCP}(\bz'|\bz)
\eqnr\label{eq:mixture}\end{align*}
where we use $\rho=0.5$ in experiments. The mixing efficiency might be half that of the oracle solution (where the optimal parameterization is known), nonetheless when taking into account the uncertainty over the parameters, the expected efficiency of the mixture proposal can be better than a single parameterization chosen ad hoc. 

We applied a Hybrid Monte Carlo (HMC) sampler to a Dynamic Bayesian Network (DBN) with nonlinear transition probabilities with the same structure as the illustrative model in figure~\ref{fig:auxformexample}. The prior and conditional probabilities are:
$\bz_1 \sim \mathcal{N}(0,\mathbf{I})$, 
$\bz_t | \bz_{t-1} \sim \mathcal{N}(tanh(\bb{W}_z \bz_{t-1} + \bb{b}_z), \sigma_z^2 \mathbf{I} )$
and $\bx_t | \bz_t \sim \text{Bernoulli}(sigmoid(\bb{W}_x \bz_{t-1}))$. The parameters were intialized randomly by sampling from $\mathcal{N}(0, \bb{I})$.
Based on the derived limiting behaviour (see table ~\ref{table:limiting_behaviour}, we can expect that such a network in CP can have very large posterior correlations if the variance of the latent variables $\sigma^2_z$ is very small, resulting in slow sampling.

To validate this result, we performed HMC inference with different values of $\sigma^2_z$, sampling the latent variables while holding the parameters fixed. For HMC we used 10 leapfrog steps per sample, and the stepsize was automatically adjusted while sampling to obtain a HMC acceptance rate of around 0.9.  At each sampling run, the first 1000 HMC samples were thrown away (burn-in); the subsequent 4000 HMC samples were kept. To estimate the efficiency of sampling, we computed the effective sample size (ESS); see e.g. \cite{kass1998markov} for a discussion on ESS.

\textbf{Results.} See table~\ref{table:ess_logz} and figure~\ref{fig:autocorr} for results. It is clear that the choice of parameterization has a large effect on posterior dependencies and the efficiency of inference. Sampling was very inefficient for small values of $\sigma_z$ in the CP, which can be understood from the limiting behaviour in table~\ref{table:limiting_behaviour}.

\subsection{Generative multilayer neural net}\label{sec:gennn}

As explained in section~\ref{sec:mclikelihood}, a hierarchical model in DNCP form can be learned using a MC likelihood estimator which can be differentiated and optimized w.r.t. the parameters $\bT$. We compare this Maximum Monte Carlo Likelihood (MMCL) method with the MCEM method for learning the parameters of a 4-layer hierarchical model of the MNIST dataset, where $\bx | \bz_3 \sim \text{Bernoulli}(sigmoid(\bb{W}_x \bz_3 + \bb{b}_x))$ and $\bz_t | \bz_{t-1} \sim \mathcal{N}(\tanh(\bb{W}_i \bz_{t-1} + \bb{b}_i), \sigma^2_{z_t} \bb{I})$. 
For MCEM, we used HMC with 10 leapfrog steps followed by a weight update using Adagrad~\cite{duchi2010adaptive}. For MMCL, we used $L \in \{10, 100, 500 \}$.
We observed that DNCP was a better parameterization than CP in this case, in terms of fast mixing. However, even in the DNCP, HMC mixed very slowly when the dimensionality of latent space become too high. For this reason, $\bz_1$ and $\bz_2$ were given a dimensionality of 3, while $\bz_3$ was 100-dimensional but noiseless ($\sigma^2_{z_1} = 0$) such that only $\bz_3$ and $\bz_2$ are random variables that require posterior inference by sampling. The model was trained on a small (1000 datapoints) and large (50000 datapoints) version of the MNIST dataset. 

\begin{figure}[t]
\includegraphics[width=8cm]{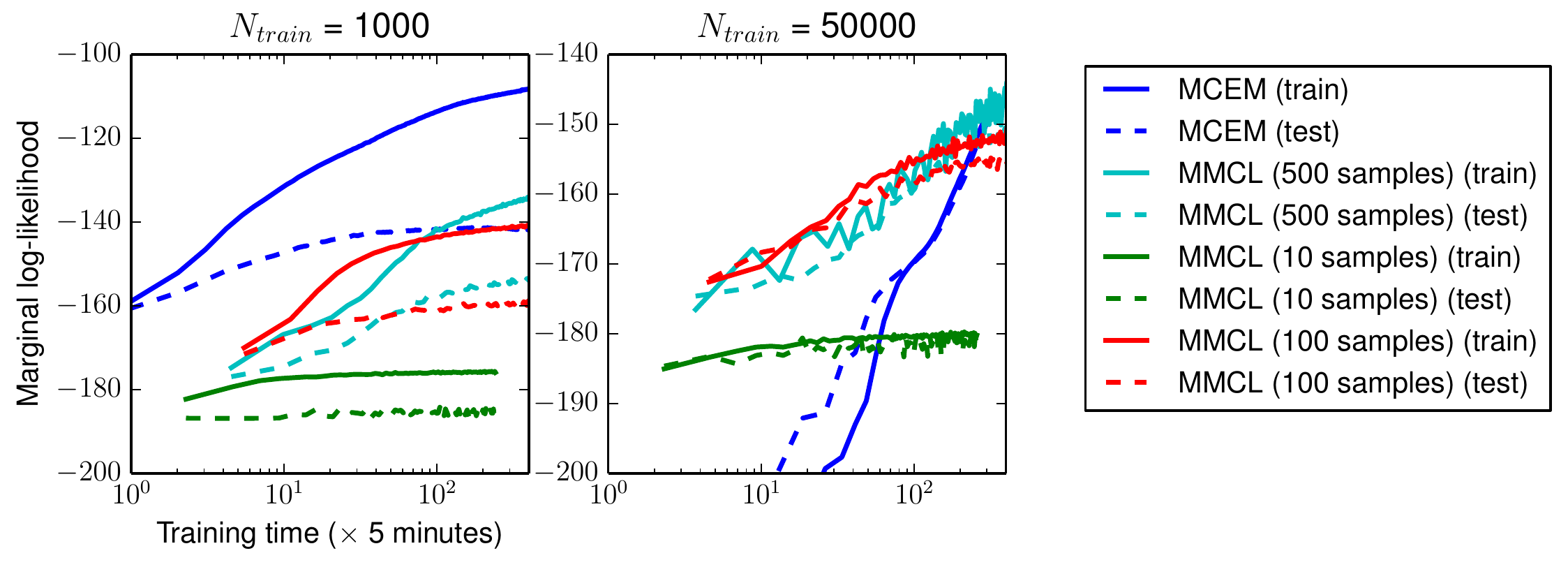}
\caption{Performance of MMCL versus MCEM in terms of the marginal likelihood, when learning the parameters of a generative multilayer neural network (see section~\ref{sec:gennn}).
}
\label{fig:mnist}
\end{figure}

\textbf{Results.} We compared train- and testset marginal likelihood. See figure~\ref{fig:mnist} for experimental results. As was expected, MCEM attains asymptotically better results. However, despite its simplicity, the on-line nature of MMCL means it scales better to large datasets, and (contrary to MCEM) is trivial to implement.

\section{Conclusion}
We have shown how Bayesian networks with continuous latent variables and generative neural networks are related through two different parameterizations of the latent variables: CP and DNCP.  A key result is that the differentiable non-centered parameterization (DNCP) of a latent variable is preferred, in terms of its effect on decreased posterior correlations, when the variable is more strongly linked to its parents than its children. Through theoretical analysis we have also shown that the two parameterizations are complementary to each other: when posterior correlations are large in one form, they are small in the other. We have also illustrated that this theoretical result can be exploited in practice by designing a MCMC strategy that mixes between both parameterizations, making it robust to situations where MCMC can otherwise be inefficient.

\section*{Acknowledgments} 

The authors thank the reviewers for their excellent feedback and Joris Mooij, Ted Meeds and Taco Cohen for invaluable discussions and input. 
\newpage
\small
\bibliography{../shared/bib}
\bibliographystyle{icml2014}

\end{document}


\title{Parameterizing Bayesian Networks as Generative Neural Networks\\Supplemental material}
\date{October 2013}
\maketitle

\section{Prerequisites}

\subsection{Computing the (squared) correlation of two jointly Gaussian distributed variables from second-order derivatives}

The covariance $C$ between two jointly Gaussian distributed variables $A$ and $B$ equals the negative inverse of the Hessian matrix of the log-joint PDF:
\begin{align*}
C 
= \left( \begin{matrix}\sigma_A^2&\sigma_{AB}^2\\ \sigma_{AB}^2 & \sigma_{B}^2 \end{matrix} \right)
= - \bb{H}^{-1} 
= \frac{1}{det(\bb{H})} \left( \begin{matrix}-H_{A}&H_{AB}\\H_{AB}&-H_{B}\end{matrix} \right)
\end{align*}

The correlation $\rho$ between any two variables $A$ and $B$ is given by: $\rho = \sigma_{AB}^2 / (\sigma_A \sigma_B)$. Using the equation above,  it can be computed from the elements of the Hessian:

\begin{align*}
\rho^2 
&= (\sigma_{AB}^2)^2 / ( \sigma_A^2 \sigma_B^2 ) \\
&= (H_{AB}/det(\bb{H}))^2 / ((-H_{A}/det(\bb{H}))(-H_{B}/det(\bb{H})) \\
&= H_{AB}^2 / (H_{A} H_{B})
\eqnr \label{eq:correlation}
\end{align*}

\subsection{Hessian}

Important to note is that derivatives of the log-posterior w.r.t. the latent variables are equal to the derivatives of log-joint w.r.t. the latent variables, therefore, 
\begin{align*}
\bb{H} = \nabla_{\bz} \nabla_{\bz}^T \log \pT(\bz|\bx) = \nabla_{\bz} \nabla_{\bz}^T \log \pT(\bx,\bz)
\end{align*}

\section{When is the proposed reparameterization beneficial for inference?}

What is the effect of the proposed reparameterization on the efficiency of inference? 

If a latent variable has a Gaussian conditional distribution, we can use the metric of squared correlation. If after reparameterization the squared correlation between a latent variable and its children will be lower, than in general this will also result in faster gradient-based inference.

For non-Gaussian posterior distributions, the log-PDF can be locally approximated as a Gaussian using a second-order Taylor expansion. Therefore, the results derived for the Gaussian case will also apply to the non-Gaussian case. The correlation computed using this approximation has an interpretation as a local dependency between the two variables.

First we will derive expressions for the squared correlations, and subsequently show how they are related.

\subsection{Notation}
Denote by $z$ the latent variable we're going to reparameterize, and by $\by$ its parents, where $y_i$ is one of the parents. The log-PDF of the corresponding conditional distribution is
\begin{align*}
\log \pT(z|\by) = \mathcal{N}(z|\bw^T \by + b, \sigma^2) = (z - \bw^T \by - b)^2/(2\sigma^2) + C
\end{align*}
The following notation is used:
\begin{align*}
\bb{H} &= \text{the Hessian of $\log \pT(\bx, \bz)$}\\
L &= \log \pT(\bx,\bz) &\text{(the log-joint, sum of all factors))} \\
\Lz &= \log \pT(z|\by) &\text{($z$'s factor, the conditional log-PDF of $z$)}\\
\Lnz &= L - \Lz &\text{(the log-joint minus $z$'s factor)}\\
\Lcz &= \text{the factors of $z$'s children}\\
\alpha &= \spd{\Lnz}{y_i}{y_i}\\
\beta &= \spd{\Lcz}{z}{z}\\
\end{align*}

\subsection{Squared Correlations}
\subsubsection{Original case}
The relevant Hessian elements are as follows:
\begin{align*}
H_{y_i y_i} &= \spd{L}{y_i}{y_i} = \alpha + \spd{\Lz}{y_i}{y_i} = \alpha - w_i^2/ \sigma^{2} \\
H_{zz} &= \spd{L}{z}{z} = \beta + \spd{\Lz}{z}{z} = \beta - 1/\sigma^2 \\
H_{y_i z} &= \spd{L}{y_i}{z} = \spd{\Lz}{y_i}{z} = w_i/\sigma^2 \\
H_{y_i y_j} &= \spd{L}{y_i}{y_j} = \spd{\Lnz}{y_i}{y_j} + \spd{\Lz}{y_i}{y_j} = \spd{\Lnz}{y_i}{y_j} + w_i/\sigma^2 \\
\end{align*}
The squared correlation between $y_i$ and $z$ is therefore:
\begin{align*}
\rho^2_{y_i,z} = \frac{(H_{y_i z})^2 }{H_{y_i y_i} H_{zz}}
= \frac{w_i^2/\sigma^4}{(\alpha - w_i^2/ \sigma^{2})(\beta - 1/\sigma^2)}
\end{align*}

\subsubsection{Reparameterized case}

We reparameterize $z$ using auxiliary variable $\ep$ with $z = g(.) = (\bw^T \by + b) + \sigma \ep$ where $\ep \sim \mathcal{N}(0,1)$.
First, we can confirm that this choice of $p(\ep)$ and $g(.)$ is correct:
\begin{align*}
\pT(z | \by) \cdot \left| \fpd{z}{\ep} \right|
&= \pT(z = g(.) | \by) \cdot \left| \fpd{z}{\ep} \right| \\
&= \mathcal{N}(\bw^T\by + b + \sigma \ep | \bw^T \by + b, \sigma^2) \cdot \sigma_z \\
&= \exp \left(\frac{((\bw^T \by + b + \sigma \ep) - (\bw^T \by + b))^2}{2 \sigma^2}\right) / \sqrt{2 \pi \sigma^2} \cdot \sigma \\
&= \exp(\ep^2/2) / \sqrt{2 \pi} = \mathcal{N}(0, 1) \\
&= p(\ep) 
\end{align*}
The Hessian elements are (using the same shorthands as above):
\begin{align*}
H_{y_i y_i} &= \spd{L}{y_i}{y_i} = \alpha + \fpd{}{y_i} \fpd{\Lcz}{y_i} = \alpha + \fpd{}{y_i} \left( w_i \fpd{\Lcz}{z} \right)  =  \alpha  + w_i^2 \beta \\
H_{\ep \ep} &= \spd{L}{\ep}{\ep} = \spd{\Lcz}{\ep}{\ep} + \spd{\log p(\beps)}{\ep}{\ep} = \sigma^2 \beta - 1\\
H_{y_i \ep} &= \spd{L}{y_i}{\ep} = \sigma w_i \beta
\end{align*}
The squared correlation between $y_i$ and $\ep$ is therefore:
\begin{align*}
\rho^2_{y_i,\ep} 
= \frac{(H_{y_i \ep})^2 }{H_{y_i y_i} H_{\ep \ep}} 
&= \frac{\sigma^2 w_i^2 \beta^2}{(\alpha  + w_i^2 \beta)(\sigma^2 \beta - 1)}\end{align*}

\subsection{Correlation inequality}
We can now compare the squared correlation, between $z$ and some parent $y_i$, before and after the reparameterization.
\begin{align*}
\rho^2_{y_i,z} &> \rho^2_{y_i,\ep} \\ 
\frac{w_i^2/\sigma^4}{(\alpha - w_i^2/ \sigma^{2})(\beta - 1/\sigma^2)}
&>
\frac{\sigma^2 w_i^2 \beta^2}{(\alpha  + w_i^2 \beta)(\sigma^2 \beta - 1)}
&\text{(derived above)}
\\
\frac{w_i^2/\sigma^4}{(\alpha - w_i^2/ \sigma^{2})(\beta - 1/\sigma^2)}
&>
\frac{w_i^2 \beta^2}{(\alpha  + w_i^2 \beta)(\beta - 1/\sigma^2)}
&\text{(re-arranging terms)}
\\
\frac{1/\sigma^4}{(\alpha - w_i^2/ \sigma^{2})}
&>
\frac{\beta^2}{(\alpha  + w_i^2 \beta)}
&\text{(terms cancelling out)}
\\
\sigma^{-2} &> \beta &\text{$\bigg($where $\beta = \spd{\Lcz}{z}{z}\bigg)$}
\\
\end{align*}
Thus we have shown the surprising fact that the correlation inequality takes on an extremely simple form where the parent-dependent values $\alpha$ and $w_i$ play no role; the inequality only depends on two properties of $z$: the relative strenghts of $\sigma$ (its noisiness) and $\beta$ (its influence on children's factors).
Informally speaking, if the noisiness of $z$'s conditional distribution is large enough compared to other factors' dependencies on $z$, then the reparameterized form is beneficial for inference.

\subsection{Limiting behaviour}

\begin{tabular}[c]{l || c | c | c | c | c | c | c | c |}
$$ & 
$\lim_{\sigma \to 0}$ &
$\lim_{\sigma \to +\infty}$ &
$\lim_{\beta \to 0}$ &
$\lim_{\beta \to +\infty}$ &
$\lim_{\alpha \to 0}$ &
$\lim_{\alpha \to +\infty}$ &
$\lim_{w_i \to 0}$ &
$\lim_{w_i \to +\infty}$ 
\\
\hline
$\rho^2_{y_i,z}$ (Original) &
1 & 
0 &
$\frac{w_i^2}{-\alpha \sigma^2 + w_i^2}$ & 
0 &
$\frac{1}{1 - \beta \sigma^2}$ & 
0 & 0 & 0\\
$\rho^2_{y_i,e}$ (Reparameterized) &
0 &
$\frac{\beta w_i^2}{\alpha + \beta w_i^2}$ &
0 &
1 &
$\frac{\beta \sigma^2}{-1 + \beta \sigma^2}$ &
0 &
0 &
$1 + \frac{1}{\beta \sigma^2 - 1}$
\\
\end{tabular}

\section{Example: Simple Linear Dynamical System}

In the case of a linear dynamical system (LDS), the log-posterior (of both the original and reparameterized form) is quadratic. This quadratic form is convenient since we can analyze and compare correlations in the Gaussian posterior distributions.

\subsection{Original form}

Take a simple model with scalar latent variables $z_1$ and $z_2$, and scalar observed variables $x_1$ and $x_2$. The joint PDF is defined as $p(x_1, x_2, z_1, z_2) = p(z_1) p(x_1|z_1) p(z_2|z_1) p(x_2|z_2)$, where $p(z_1) = \mathcal{N}(0,1)$, $p(x_1|z_1)=\mathcal{N}(z_1,\sigma_x^2)$, $p(z_2|z_1)=\mathcal{N}(z_1,\sigma_z^2)$ and $p(x_2|z_2)=\mathcal{N}(z_2,\sigma_x^2)$. Note that the parameter $\sigma_z$ controls sets the dependency between the latent variables, and $\sigma_x$ the dependency between latent and observed variables. The joint log-PDF is: $\log p(z_1, z_2 | x_1, x_2) = - z_1^2/2 - (x_1 - z_1)^2/(2\sigma_x^2) - (z_2-z_1)/(2\sigma_z^2) - (x_2 - z_2)^2/(2\sigma_x^2) + C$, where $C$ is constant w.r.t. $z_1$ and $z_2$. Recall that the posterior log-PDF is equal to the joint log-PDF, up to a constant.

Let $\bb{H} = \bigl( \begin{smallmatrix}H_{11}&H_{12}\\H_{21}&H_{22}\end{smallmatrix} \bigr)$ and $L=\log p(z_1,z_2|x_1,x_2)$, then:
\begin{align*}
H_{11} &= \spd{L}{z_1}{z_1} = -1 - \frac{1}{\sigma_x^2} - \frac{1}{\sigma_z^2}\\
H_{22} &= \spd{L}{z_2}{z_2} = - \frac{1}{\sigma_x^2} - \frac{1}{\sigma_z^2}\\
H_{21} &= \spd{L}{z_1}{z_2} = \frac{1}{\sigma_z^2}
\end{align*}

From eq.~\eqref{eq:correlation}, the squared correlation between the two variables is now:
\begin{align*}
\rho^2 &= H_{12}^2 / (H_{11} H_{22})\\
&= \frac{1/ \sigma_z^4}{(1 + 1/\sigma_x^2 + 1/\sigma_z^2)\cdot(1/\sigma_x^2 + 1/\sigma_z^2)}
\end{align*}

\subsection{Reparameterized form}

Now we reparameterize $z_2$ such that it is conditionally deterministic given a new auxiliary variable $\ep_2$. We will do this by choosing $p(\ep_2) = \mathcal{N}(0,1)$ and $z_2 = g_2(z_1, \ep_2, \sigma_z) = z_1 + \sigma_z \cdot \ep_2$, and consequently $\ep_2 = g^{-1}(z_1, z_2, \sigma_z) = (z_2 - z_1)/\sigma_z$. 

First, we can confirm that this choice of $p(E_2)$ and $g_2(.)$ is correct:
\begin{align*}
p(\ep_2) 
&= p(z_2 = g(z_1, \ep_2, \sigma_z) | z_1) \cdot \left| \fpd{z_2}{\ep_2} \right| \\
&= \mathcal{N}(z = z_1 + \sigma_z \ep_2, \sigma_z^2) \cdot \sigma_z \\
&= \exp \left(\frac{((z_1 + \sigma_z \ep_2) - z_1)^2}{2 \sigma_z^2}\right) / \sqrt{2 \pi \sigma_z^2} \cdot \sigma_z \\
&= \exp(\ep_2^2/2) / \sqrt{2 \pi} \\
&= \mathcal{N}(0, 1)
\end{align*}

For sake of notational clarity and completeness, $z_1$ will be trivially parameterized as $z_1 = e_1$. After reparameterization: $\log p(x_2|z_2) = - (x_2 - z_2)/(2\sigma_x^2) + C = - (x_2 - (e_1 + \sigma_z e_2))/(2\sigma_x^2) + C = \log p(x_2 | e_1, e_2)$, where $C$ is constant.  The full auxiliary joint PDF is: $\log p(e_1, e_2 | x_1, x_2) = - e_1^2/2 - e_2^2/2 - (x_1 - e_1)^2/(2\sigma_x^2) - (x_2 - (e_1 + \sigma_z e_2))^2/(2\sigma_x^2) + C$, where $C$ is constant w.r.t. the variables.

Let $\bb{H}' = \bigl( \begin{smallmatrix}H'_{11}&H'_{12}\\H'_{21}&H'_{22}\end{smallmatrix} \bigr)$ and $L'=\log p(e_1,e_2|x_1,x_2)$, then:
\begin{align*}
H'_{11} &= \spd{L'}{z_1}{z_1} = -1 - \frac{2}{\sigma_x^2}\\
H'_{22} &= \spd{L'}{z_2}{z_2} = -1 - \frac{\sigma_z^2}{\sigma_x^2}\\
H'_{21} &= \spd{L'}{z_1}{z_2} = \frac{\sigma_z}{\sigma_x^2}
\end{align*}

From eq.~\eqref{eq:correlation}, the squared correlation between the two variables is now:
\begin{align*}
\rho'^2 &= H'_{12} / (| H'_{11} | | H'_{22} |)\\
&= \frac{ \sigma_z^2/ \sigma_x^4}{(1 + 2/\sigma_x^2)\cdot(1 + \sigma_z^2/\sigma_x^2)}
\end{align*}

\subsection{Comparison}

For what choice of parameters does the reparameterized form yield smaller correlation?
\begin{align*}
\rho^2 &> \rho'^2 \\
\frac{1/ \sigma_z^4}{(1 + 1/\sigma_x^2 + 1/\sigma_z^2)\cdot(1/\sigma_x^2 + 1/\sigma_z^2)} &> \frac{ \sigma_z^2/ \sigma_x^4}{(1 + 2/\sigma_x^2)\cdot(1 + \sigma_z^2/\sigma_x^2)} \\
\sigma_z &< \sigma_x
\end{align*}

\section{General equations for Hessian elements}

Denote by $\bb{H}$ the Hessian of the joint PDF. The original diagonal Hessian elements are as follows:
\begin{align}
H_{jj} = \spd{\Lxz}{z_j}{z_j} &= \spd{\Lzj}{z_j}{z_j} + \spd{\LpajOrig}{z_j}{z_j}
\label{HjjOrig}
\end{align}
where $\Lxz = \log \pT(\bx,\bz)$ is the original joint PDF, $\Lzj = \log \pT(\bz_j|\bpa_j)$ is the conditional PDF of $\bz_j$, and $\LpajOrig$ is the sum of the conditional log-PDFs of the children of $Z_j$  before the reparameterization. The off-diagonal elements are determined by parent-child dependencies. If some variable $Z_i$ is a parent of $Z_j$, then:
\begin{align}
H_{ij} = \spd{\Lxz}{z_i}{z_j} &= \spd{\Lzj}{z_i}{z_j}
\label{HijOrig} 
\end{align}

Denote by $\bb{H}'$ the Hessian of the auxiliary posterior, where each $Z_j \in \bb{Z}$ is reparameterized using $E_j$, i.e. $\bb{H}' = \nabla_{\beps} \nabla_{\beps}^T \log \pT(\beps|\bx) = \nabla_{\beps} \nabla_{\beps}^T \log \pT(\bx,\beps)$. The diagonal elements are as follows:
\begin{align*}
\bb{H}'_{jj} &= \spd{\Lxe}{e_j}{e_j} = \spd{\Lej}{e_j}{e_j} + \fpd{}{e_j} \left( \fpd{\LpajAux}{z_j} \fpd{z_j}{e_j} \right) \\
&= \spd{\Lej}{e_j}{e_j} + \fpd{z_j}{e_j} \fpd{}{e_j} \fpd{\LpajAux}{z_j} + \fpd{\LpajAux}{z_j} \fpd{}{e_j} \fpd{z_j}{e_j} &\text{(product rule)} \\
&= \spd{\Lej}{e_j}{e_j} + \left(\fpd{z_j}{e_j}\right)^2 \spd{\LpajAux}{z_j}{z_j} + \spd{z_j}{e_j}{e_j} \fpd{\LpajAux}{z_j} &\text{(rearranging terms)} \\
\eqnr\label{eq:HjjAux}\end{align*}
If some variable $Z_i$ is a parent of $Z_j$ then after reparameterization:
\begin{align*}
\bb{H}'_{ij} &= \spd{\Lxe}{e_i}{e_j} = \fpd{z_i}{e_i} \left( \fpd{}{z_i} \fpd{\LpajAux}{z_j} \fpd{z_j}{e_j} \right) &\text{($e_j$ influences $L$ through $z_j$)} \\
&= \fpd{z_i}{e_i} \left( \fpd{z_j}{e_j} \fpd{}{z_i} \fpd{\LpajAux}{z_j} + \fpd{\LpajAux}{z_j} \fpd{}{z_i} \fpd{z_j}{e_j} \right) &\text{(product rule)}\\
&= \fpd{z_i}{e_i} \left( \fpd{z_j}{e_j} \spd{\LpajAux}{z_j}{z_j} \fpd{z_j}{z_i} + \fpd{\LpajAux}{z_j} \spd{z_j}{z_i}{e_j} \right) \\
&= \fpd{z_i}{e_i} \left( \fpd{z_j}{e_j} \fpd{z_j}{z_i} \spd{\LpajAux}{z_j}{z_j} + \spd{z_j}{z_i}{e_j} \fpd{\LpajAux}{z_j} \right) &\text{(rearranging terms)}\\
\eqnr
\label{eq:HijAux}
\end{align*}
where $\Lxe = \log \pT(\bx,\beps)$ is the auxiliary joint PDF and $\LpajAux$ is the sum of conditional PDFs of the observed variables after parameterization.

\bibliography{bib}
\bibliographystyle{icml2013}